\documentclass[11pt,a4paper]{article}
\usepackage[hyperref]{acl2019}
\usepackage{times}
\usepackage{latexsym}

\usepackage{url}

\usepackage{amsmath, amsthm, amssymb}    
\usepackage{amsthm}
\usepackage{graphicx}
\usepackage{amsmath}
\usepackage{amsfonts}
\usepackage{graphicx}
\usepackage{verbatim}
\usepackage{booktabs}
\usepackage{subfig}
\usepackage{amsthm}
\usepackage{amssymb}
\usepackage{qtree}
\usepackage{graphicx}
\usepackage{caption}
\usepackage{multirow}
\usepackage{mathrsfs}
\usepackage{tablefootnote}
\usepackage{arydshln}
\usepackage{lscape}
\usepackage{afterpage}
\usepackage{todonotes}
\usepackage{arydshln}
\usepackage{color}
\usepackage[linewidth=1pt]{mdframed}

\newtheorem{definition}{Definition}

\usepackage{url}
\usepackage{caption}
\usepackage{xcolor}
\usepackage{calc}

\usepackage{pgfplots} 
  
\aclfinalcopy 

\title{Distant Learning for Entity Linking with Automatic Noise Detection}
\author{Phong Le$^1$ \and Ivan Titov$^{1,2}$ \\
  $^{1}$University of Edinburgh   $\;^{2}$University of Amsterdam \\
  {\tt lephong.xyz@gmail.com $\;\;$ ititov@inf.ed.ac.uk}
  \\}

\date{}

\begin{document}
\maketitle

\begin{abstract}

Accurate entity linkers have been produced for domains and languages where annotated data (i.e., texts linked to a knowledge base) is available. However, little progress has been made for the settings  where no or very limited amounts of labeled data are present (e.g., legal or most scientific domains). 
In this work, we show how we can learn to link mentions without having any labeled examples, only a knowledge base and a collection of unannotated texts from the corresponding domain. In order to achieve this, we frame the task as a multi-instance learning problem and rely on surface matching to create initial noisy labels. As the learning signal is weak and our surrogate labels are noisy, we introduce a noise detection component in our model: it lets the model detect and disregard examples which are likely to be noisy.
Our method,  jointly learning to detect noise and link entities, greatly outperforms the surface  matching baseline. For a subset of entity categories, it even approaches the performance of supervised learning.


\end{abstract}

\section{Introduction}
\label{sec:intro}

Entity linking (EL) is the task of linking potentially ambiguous textual mentions to the corresponding entities in a knowledge base.
Accurate entity linking is crucial in many natural language processing tasks, including information extraction 
\cite{hoffart-EtAl:2011:EMNLP} and
question answering 
\cite{yih-EtAl:2015:ACL-IJCNLP}.
Though there has been significant progress in entity linking recently ~\cite{P11-1138, hoffart-EtAl:2011:EMNLP, TACL494, globerson-EtAl:2016:P16-1, TACL1065, ganea-hofmann:2017:EMNLP2017, P18-1148}, previous work has focused on supervised learning.  Annotated data necessary for supervised learning is available for certain knowledge bases and domains. For example, one can directly use web-pages linking to Wikipedia to learn a Wikipedia linker. Similarly, there exist domain-specific sets of manually annotated documents  (e.g., AIDA-CoNLL news dataset for YAGO~\cite{hoffart-EtAl:2011:EMNLP}). However, for many ontologies and domains annotation is not available or limited (e.g., law).
Our goal is to develop a method which does not rely on any training data besides unlabeled texts and a knowledge base.

\begin{figure*}
    \centering
    \includegraphics[width=0.9\textwidth]{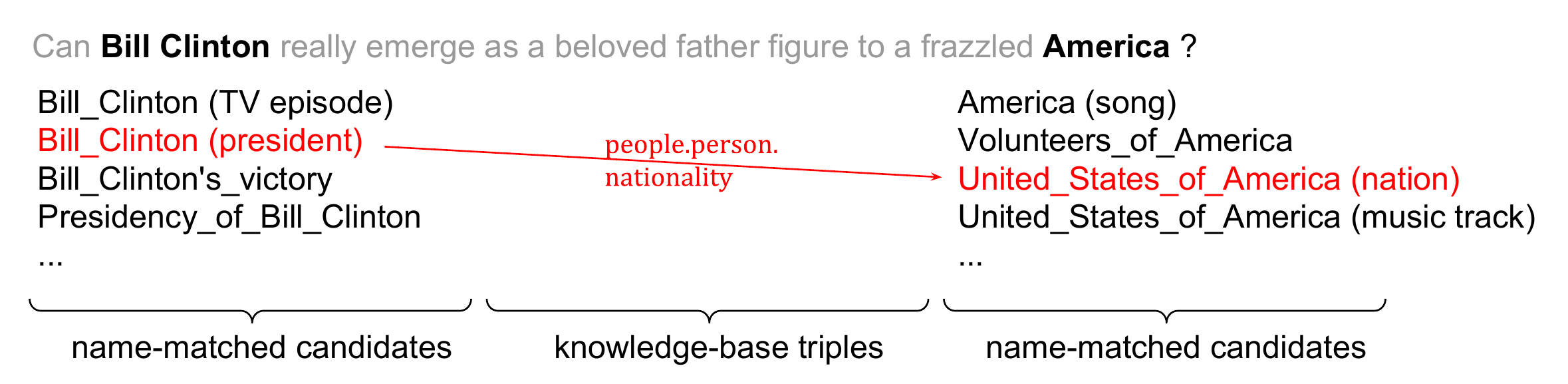}
    \caption{We annotate raw sentences using entity names and 
    knowledge base triples. In training, we keep only red entities as 
    positive candidates. 
    In testing, we consider $|E^+|=100$ name-matched candidates.}
    \label{fig:example}
\end{figure*}

In order to construct such a method, we use an insight from simple surface matching heuristics (e.g., \newcite{riedel2010modeling}). Such heuristics choose entities from a knowledge base by measuring the overlap between the sets of content words in the mention and in the entity name. For example, in Figure~\ref{fig:example}, the entities
\textsc{Bill Clinton (president) } and \textsc{Presidency of 
Bill Clinton} both have two matching words with the mention {\it Bill Clinton}.
Whereas we will see in our experiments that this method alone is not particularly accurate at selecting the best entity, the candidate lists it provides often include the correct entity. 
This implies that we can both focus on learning to select candidates from these lists and, less obviously, that we can leverage the lists as weak or distant supervision.

We frame this distance learning (DL) task as  the multi-instance learning (MIL) problem \cite{Dietterich:1997:SMI:249678.249682}. In MIL, each bag of examples is marked with a class label: the label indicates that the bag contains at least one example corresponding to that class. Relying on such labeled bags, MIL methods aim at learning classifiers for individual examples.

Our DL problem can be regarded as a binary version of MIL. For a list of entities (and importantly given the corresponding mention and its document context), we assume that we know if the list contains a correct entity or not. The `positive lists' are essentially top candidates from the matching heuristic.
For example, the four candidate entities for the mention `Bill Clinton' in Figure ~\ref{fig:example} could be marked as a positive set. 
The `negative lists' are randomly sampled sets of entities from the knowledge base. As with other MIL approaches, while relying on labeled lists, we learn to classify individual entities, i.e. to predict if an entity should be linked to the mention. 

One important detail is that the classifier must not have access to information which and how many words match between the mention and the entity name. If it would know this, it would easily figure out which entity set is a candidate list and which one consists of randomly generated entities based solely on this information. Instead, by hiding it from the classifier, we force the classifier to extract features of the  mention and its context predictive of the entity properties (e.g., an entity type), and hence ensure generalization.

Unfortunately, our supervision is noisy. The positive lists will often miss the correct entity for the given mention. 
This confuses the MIL model.
In order to address this issue, we, jointly with the MIL model, learn a classifier which detects  potentially problematic candidate lists.
In other words, the classifier predicts how likely a given list is noisy (i.e., how much we should trust it). The probability is then used to weight the corresponding term
in the objective function of the MIL model. 
By jointly training the MIL model and the noise detection classifier, we effectively let the MIL model choose which examples to use for training. As we will see in our experimental analysis, this joint learning method leads to a substantial improvement in performance. We also confirm that the noise detection model is generally able to identify and exclude wrong candidate lists by comparing its predictions to the gold standard.

DL is the mainstream approach to learning relation extractors (RE)~\cite{mintz-EtAl:2009:ACLIJCNLP,riedel2010modeling}, a problem related to entity linking. However, the two instantiations of the DL framework are very different. For RE, a bag of sentences is assigned to a categorical label (a relation). For EL, we assign a bag of entities, conditioned on the mention, to a positive class (correct) or a negative class (incorrect).

We evaluate our approach on the news domain for English as, having gold standard annotation (AIDA CoNLL), we can both assess performance and compute the upper bound, given by supervised learning. Nevertheless, we expect that our methodology is applicable to a wider range of knowledge bases, as long as unlabeled texts can be obtained for the corresponding domain. We plan to verify this claim in future work. In addition, we restrict ourselves to sentence-level modeling and, unlike state-of-the-art supervised methods,~\cite{TACL1065, ganea-hofmann:2017:EMNLP2017, P18-1148} ignore interaction between linking decisions in the document. Again, it would be interesting to see if such global modeling would be beneficial in the distance learning setting.

Our contributions can be summarized as follows
\begin{itemize}
    \item we show how the entity linking problem can be framed as a distance learning problem, namely as a binary MIL task;
    \item we construct a model for this task;
    \item we introduce a method for detecting noise in the automatic annotation;
    \item we demonstrate the effectiveness of our approach on a standard benchmark.
\end{itemize}

\section{Entity linking as MIL} 
\label{sec:entity linking}

For each entity mention $m$ with context $c$, we denote $E^+$ and $E^-$ 
lists of positive candidates and negative candidates: $E^+$ should have a high 
chance of containing the correct entity $e$, while $E^-$ should include only incorrect entities. As standard in MIL, this will be the only supervision
the model receives at training time. When using this supervision, the model will need to learn to decide which entity  $e$ in $E^+$ is most likely to  correspond to the mention-context pair $(m, c)$.  
At test time, the model with be provided with the list $E^{+}$ and will need to select an entity from this list. 

Performing entity linking in two stages, candidate 
selection (generating candidate lists) and entity disambiguation (choosing an entity from the list), is standard in EL, with the first stage usually handled with heuristics and the second one approached with  statistical modeling~\cite{P11-1138, hoffart-EtAl:2011:EMNLP}. 

However, 
in our DL setting both stages change substantially.
The candidate selection stage relies primarily on a surface matching heuristic, as described in Section~\ref{sec:dataset}. Whereas supervised learning for the disambiguation stage (e.g., \newcite{hoffart-EtAl:2011:EMNLP}) is replaced with MIL learning as described below in Section~\ref{sec:models}.\footnote{Supervised learning is equivalent to assuming that $E^{+}$ are singletons containing only the gold-standard entity.}

To make the following sections clear, we introduce the following terms. 

\begin{definition}
A data point is a tuple $\langle m, c, E^+, E^- \rangle$ of 
mention $m$, context $c$, positive set $E^+$, and negative set $E^-$.
In testing, $E^- = \emptyset$.
\end{definition}

\begin{definition}
A data point $\langle m, c, E^+, E^- \rangle$ is \emph{noisy} 
if $E^+$
does not contain the correct entity for mention $m$. If a data point is not noisy,
we will refer to it as \emph{valid}.
\end{definition}

\section{Models}
\label{sec:models}

We introduce two approaches. The first one directly applies MIL, disregarding the fact that many data points are noisy. The second one addresses this shortcoming by integrating a noise detection component. 


\subsection{Model 1: MIL}
\label{subsec:model 1}



\paragraph{Encoding context}
Context $c$ is the entire $l$-word sentence $w_1, ..., w_l$ which also includes the mention
$m=(w_h, ..., w_k), 1 \leq h \leq k \leq l$. 
We use a BiLSTM to encode sentences.
The input to the BiLSTM is a concatenation 
$\mathbf{w}^*_i = [\mathbf{w}_i, \mathbf{p}_i]$
where $\mathbf{p}_i \in \mathbb{R}^{d_p}$ is position embedding 
and $\mathbf{w} \in \mathbb{R}^{d_w}$ is from GloVe\footnote{
\url{http://nlp.stanford.edu/data/glove.840B.300d.zip}} \cite{pennington-socher-manning:2014:EMNLP2014}.
Forward $\mathbf{f}_i$ and backward $\mathbf{b}_i$ states of BiLSTM are fed into the classifier described below.

\paragraph{Entity embeddings}
In this work, we use a simple and scalable approach which involves computing entity embeddings on the fly using associated types.
For instance, the TV episode 
\textsc{Bill Clinton} is associated with several types including 
\textsc{base.type\_ontology.non\_agent} and
\textsc{tv.tv\_series\_episode}.
Specifically, in order to produce an entity embedding,  each type $t$ is assigned a vector $\mathbf{t} \in \mathbb{R}^{d_t}$. 
We then compute a vector for entity $e$ as
\[
\mathbf{e} = \text{ReLU}(\mathbf{W}_e \frac{1}{|T_e|}\sum_{t \in T_e} \mathbf{t} + \mathbf{b}_e),
\]
where $T_e$ is the set of $e$'s types, and $\mathbf{W}_e \in \mathbb{R}^{d_e \times d_t}$, 
$\mathbf{b} \in \mathbb{R}^{d_e}$ are a weight matrix and a bias vector.

More sophisticated approaches to producing entity embeddings  (e.g., using relational graph convolutional networks~\cite{schlichtkrull2018modeling}) are likely to yield further improvements.

\paragraph{Scoring a candidate}
We use a one-hidden layer feed forward NN to compute score compatibility between a context-mention pair $(m, c)$ and an entity $e$: 
\[
g(e, m, c) = \text{FFN}_g([\mathbf{e}, \mathbf{f}_{h-1}, \mathbf{b}_{h-1}, \mathbf{f}_k, \mathbf{b}_k])
\]
If $e^*$ is the correct entity, we want $g(e^*, m, c) > g(e, m, c) $ 
for any  entity $e \ne e^*$. 

\paragraph{Training}
Recall that for each mention-context pair $(m, c)$, we have a positive set $E^+$ and 
a negative set $E^-$. 
We want to train the model to score at least one candidate in $E^+$  
 higher than any candidate in $E^-$. 
We use the max-margin loss to achieve this. Let
\begin{align*}
l(m, c) &= [ \max_{e \in E^-} g(e, m, c) + \delta - \max_{e \in E^+} g(e, m, c) ]_+ \\
L_1 &= \sum_{(m, c) \in D}  l(m, c) 
\end{align*}
where $\delta$ is a margin and $[x]_+ = x$ if $x > 0$ else $0$; 
$D$ is the training set. 
We want to minimize $L_1$ with respect to the model parameters. We rely on  Adam optimizer and employ early stopping.

\subsection{Model 2: MIL with Noise Detection (MIL-ND)}
\label{subsec:model 2}

The model 1 ignores the fact that many data points are noisy, i.e. $E^+$ may not contain the correct entity. We address this by integrating a binary noise detection (ND) classifier which predicts if a data point is noisy. Intuitively, data points classified as noisy need to be discarded from training of the EL model. In practice, we weight them with the confidence of the ND classifier.
As discussed below, we train the ND classifier jointly with the EL model.

\paragraph{Representation for $E^+$}

The ND classifier needs to decide if there is at least one entity in the list $E^+$ corresponding to the mention-context pair $(m,c)$. The question is now how to represent $E^+$ to make classification as easy as possible. One option is to use mean pooling, but this would result in uninformative representations, especially for longer candidate lists. Another option is max pooling, but it would not take into account which mention-context pair $(m,c)$ is currently considered, so also unlikely to yield informative features of $E^+$.
Instead
we use attention, with the attention weight computed as a function of $(m,c)$:
\[
\mathbf{e}_{E^+} = \sum_{e \in E^+} \alpha_e \mathbf{e}
\]
where $\alpha_e$ are attention weights
\[
\alpha_e = \frac{\exp\{g'(e, m, c)/T\}}{\sum_{e' \in E^+} \exp\{g'(e', m, c)/T\}},
\]
where $g'$ is a score function. Instead of learning
a separate attention function for the ND classifier, we reuse the one from the EL model, i.e. $g = g'$. 
This will reduce the number of parameters and make the method less prone to overfitting.
Maybe more importantly, we expect that 
the better the entity disambiguation score function is, the better 
the ND classifier is, so tying the two together may provide an appropriate inductive bias. $T$ is temperature, controlling 
how sharp $\alpha_e$ should be. We found that 
a small $T = 1/3$ stabilizes the learning. 

\paragraph{Noise detection}
We use a binary classifier to detect noisy data points.
The probability that a data point is noisy is defined as
\begin{align*}
&p_N(1|m, c, E^+) = \\
&\sigma\left(\frac{\text{FFN}_f([\mathbf{e}_{E^+}, 
\mathbf{f}_{h-1}, \mathbf{b}_{h-1}, \mathbf{f}_k, \mathbf{b}_k])}{T}\right),
\end{align*}
%
$\sigma$ is the logistic sigmoid function. For simplicity, we use the same $T$ as above.

\paragraph{Training}
Our goal is to down-weight potentially noisy data points. 
Our new loss is
\begin{align*}
L_2 =& \sum_{(m, c) \in D} p_N(0|m, c, E^+) l(m, c) + \\
&\eta \times \text{KL}(\frac{\sum_{(m,c) \in D} p_N( \cdot | m, c, E^+)}{|D|}  | p^*_N),
\end{align*}
where $p^*_N$ is a prior distribution indicating our beliefs about the proportion of noisy data points; $\eta$ is a hyper-parameter. We optimize the objective with respect to the parameters of both ND and EL models. 
The second term is necessary, as without it the loss can be trivially minimized by the ND classifier  predicting that all data points are noisy with the probability of 1. This would set the first term to exactly zero. 

Intuitively, when using the second term, the model can disregard certain data points but disregarding too many of them incurs a penalty. Which data points are likely to be disregarded? Presumably the ones less consistent with the predictions of the EL model. In other words, joint training of EL and ND models encourages learning an entity-linking scoring function consistent with a large proportion of the data set but not necessarily with the entire data set. As we will see in the experimental section, the ND classifier indeed detects noisy data points rather than chooses some random subset of the data.\footnote{The second term is similar to that used in posterior regularization~\cite{ganchev2010posterior} and generalized expectation criteria method~\cite{mann2010generalized}.}


We use the same optimization procedure as for the model 1. The second term is estimated at the mini-batch level.



\paragraph{Testing}
Differently from model 1, with model 2 we have two options on how to use it at test time: 
\begin{itemize}
\item ignoring the ND classifier, thus doing entity disambiguation the same way as for 
model 1, or 

\item using the ND classifier as 
a mechanism to decide if the test data point should be classified 
as `undecidable' or not. 
Specifically, if $p_N(1 | m, c, E^+) > \tau$, model 2 will not output 
an entity for this data point. This should increase precision, 
as at test time $E^+$ also may not contain the correct entity. 
\end{itemize}
We call the two versions MIL-ND and $\tau$MIL-ND, respectively.

\section{Dataset}
\label{sec:dataset}


We describe how we create our dataset.
We use Freebase\footnote{\url{https://developers.google.com/freebase/}. 
Freebase is chosen because it contains the largest set of entities among available knowledge bases 
\cite{DBLP:journals/corr/abs-1809-11099}.}, 
though our approach should be applicable to many other knowledge bases.
Brief statistics of the dataset are shown in Table~\ref{tab:dataset info}. 

\begin{table}
    \centering
    \begin{tabular}{l|r|r}
        Set & \# sentences & \# mentions \\
        \hline 
        Train & 170,000 & 389,989 \\
        Dev & 2,275 & 4,603 \\
        Test & 2,414 & 4,286 \\
    \end{tabular}
    \caption{The statistics of the proposed dataset.}
    \label{tab:dataset info}
\end{table}

\subsection{Training set}
\label{subsec:trainset}
We took raw texts from the New York Times corpus, 
tagged them with the CoreNLP named entity recognizer\footnote{\url{https://stanfordnlp.github.io/CoreNLP/}} 
\cite{P14-5010}.
We then selected only sentences that contain at least two entity mentions. 
We did this because on the one hand in most applications of EL we care about relations between entities (e.g., relation extraction), on the other hand, it provides us with an opportunity to prune the candidate list effectively, as discussed below. Note that we do it only for training.

For each mention $m$ we carried out candidate selection as follows.
First, we listed all entities which names contain all words of $m$.
For instance, ``America'' (Figure~\ref{fig:example}) can be both the nation \textsc{United States of America} and
Simon \& Garfunkel's song \textsc{America}. 
We ranked these chosen entities by the entity ordering in the knowledge base 
(i.e., the one that appears first in the knowledge base would be ranked first); 
for Freebase this order is correlated with prominence.

Second, for each mention (e.g., ``Bill Clinton''), we kept only entities
which participate in a relation with one of the candidate entities for another mention in the sentence.
For example, \textsc{Bill Clinton (president)} is kept because it is in the \textsc{person.person.nationality} relation 
with the entity \textsc{United States of America (nation)}.

Last, to keep candidate lists manageable, we selected only $|E^+|=100$ 
candidates from step 2 for each mention as positive candidates.
During training, we sampled $|E^-|=10$ candidates from the rest of the knowledge base
as negative candidates.

\subsection{Development and test sets}
\label{subsec:devtest}
We took manually annotated AIDA-A and AIDA-B as development and test sets
\cite{hoffart-EtAl:2011:EMNLP}.
We turned the ground truth Wikipedia links
in these sets to 
Freebase entities, thanks to the mapping available in Freebase.\footnote{We could not handle
NIL cases here because the  knowledge base used to annotate AIDA-CoNLL 
is different from the one we use.}

\begin{figure}
    \includegraphics[width=0.48\textwidth]{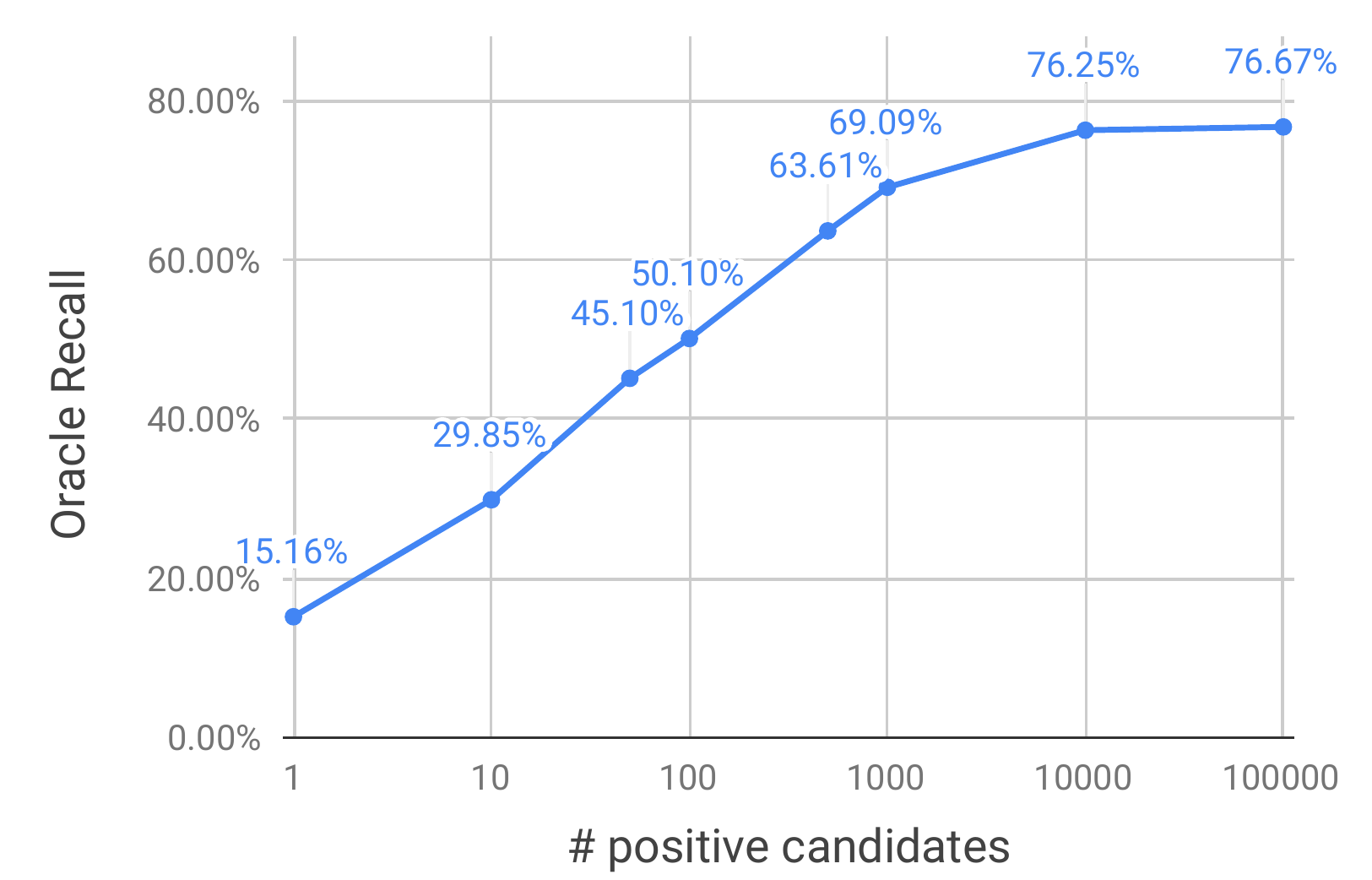}
    \caption{Oracle recall as a function of $|E^+|$ (the number of positive candidates)
    on the development set.}
    \label{fig:oracle recall}
\end{figure}

Candidate selection was done in the same way as for training, except for not filtering out sentences with only 1 entity (i.e. no step 2 from Section~\ref{subsec:trainset}).
The oracle recall for surface name matching (i.e. step 1 from Section~\ref{subsec:trainset}) is 77\%.
It goes down to 50\% if we restrict $|E^+| = 100$
(see Figure~\ref{fig:oracle recall}). We believe that there are straightforward ways to improve
the selection heuristic (e.g., modifying the string matching heuristic or using word embeddings to match words in entity names and words in the mention) but 
we leave this for future work.

Note that because AIDA CoNLL dataset is based on 
Reuters newswire articles, these development and test sets do not overlap with the training set.

\section{Experiments}
We evaluated the models above using the data from Section~\ref{sec:dataset}. 
The source code and the data are available at 
\url{https://github.com/lephong/dl4el}

We ran each model five times and report mean and 95\% confidence interval of
three metrics: (micro) precision, (micro) recall, and (micro) F1
\cite{cornolti2013framework} under two settings:
\begin{itemize}
    \item `All': all mentions are taken into account, 
    \item `In $E^+$': only mentions with $E^+$ containing the correct entity are considered.
\end{itemize}
The latter, though not realistic, is interesting as it lets us concentrate on the contribution of the disambiguation model, and ignore cases which are hopeless with the considered candidate selection method.

Note that, for system outputting exactly one entity for each mention (e.g., MIL model 1), 
precision and recall are equal.

\subsection{Systems}

We compared our models against `Name matching'.
It was proposed by \newcite{riedel2010modeling} for RE: 
a mention is linked to an entity if it matches the entity's name. 
For tie cases, we chose the first matched entity appearing in Freebase.
For instance, ``America'' is linked to the song instead of the nation. 
To our knowledge, name matching is the only method tried in previous work for our 
setting (i.e. with no annotated texts). 

We also compared with a supervised version of model 1. 
We used the same method in Section~\ref{subsec:devtest}
to convert AIDA CoNLL training set, 
with $E^+$ being singletons consisting of the correct entity provided by human annotators.
This system can be considered 
as an upper-bound of our two models because: (i) it is trained in supervised rather than MIL setting with gold standard labels rather than weak supervision, and (ii) 
the training set is in the same domain (i.e. Reuter) with the test set.
Although it uses only entity types but no other entity-related 
information for entity disambiguation, 
in Appendix B we show that this system 
performs on par with \newcite{hoffart-EtAl:2011:EMNLP} when evaluated in their setting.

Note that comparison with supervised linkers proposed in previous work 
is not possible as they require Wikipedia 
(see Section~\ref{sec:related work}) 
for candidate selection, as a source of supervision, 
and often for learning entity embeddings.

We tuned  hyper-parameters  on the 
development set. Details are in Appendix A.
Note that, in model 2 (both MIL-ND and $\tau$MIL-ND), 
we set the prior $p^*_N(1)$ to 0.9, i.e. requiring 90\% of 
training data points should be ignored.\footnote{
Section~\ref{subsec:analysis} shows that 90\% is too high, 
but it helps the model to rely only on those entity disambiguation decisions
that are very certain.}
We experimented with $|E^+| = 100$ for both training and testing. 
For training, we set $|E^-| = 10$.

\begin{table*}[!ht]
\centering
    \begin{tabular}{l|ccc|ccc}
         & \multicolumn{3}{c|}{All} & \multicolumn{3}{c}{In $E^+$} \\
        \hline
        System & P & R & F1 & P & R & F1 \\
        \hline
        Name matching & 15.03 & 15.03 & 15.03 & 29.13 & 29.13 & 29.13 \\
        MIL (model 1) & 35.87 & 35.87 & 35.87 $\pm 0.72$ & 
                        69.38 & 69.38 & 69.38 $\pm 1.29$ \\
        MIL-ND (model 2)& 37.42 & \textbf{37.42} & 37.42 $\pm 0.35$ & 
                          72.50 & \textbf{72.50} & \textbf{72.50} $\pm 0.68$ \\
        $\tau$MIL-ND (model 2)& \textbf{38.91} & 36.73 & \textbf{37.78} $\pm 0.26$ & 
                                \textbf{73.19} & 71.15 & 72.16 $\pm 0.48$ \\
        \hline
        Supervised learning & 42.90 & 42.90 & 42.90 $\pm 0.59$ & 
                              83.12 & 83.12 & 83.12 $\pm 1.15$ \\
    \end{tabular}
    \caption{Results on the test set under two settings. 
    95\% confidence intervals of F1 scores are shown.}
    \label{tab:results}
\end{table*}

\subsection{Results}
Table~\ref{tab:results} shows results on the test set. 
`Name matching' is far behind the two models.
Many entities in the knowledge base have similar or even identical names, so relying only on the surface form does not result in an effective method.\footnote{For instance, there are 36 entities named \textsc{Bill Clinton}, 
and 248 entities having `Bill Clinton' in their name.} 

MIL-ND achieves higher precision, recall, and F1 than 
MIL, this suggests that the ND classifier helped to eliminate bad data points 
during training. Using its confidence at test time ($\tau$MIL-ND,  `All' setting) was also beneficial in terms of precision and F1 (it cannot possibly increase recall).
Because all the test data points are valid for the `In $E^+$' setting, 
using the ND classifier had a slight negative effect on F1.

MIL-ND significantly outperforms MIL: the 95\% confidence intervals for them do not overlap.
However, this is not the case for MIL-ND and $\tau$MIL-ND. 
We therefore conclude that the ND classifier is clearly helpful for training
and potentially for testing.


\subsection{Analysis}
\label{subsec:analysis}

\paragraph{Error types}
In Table~\ref{tab:ner} we classified errors according to named entity types thanks to the 
annotation from \newcite{TjongKimSang-DeMeulder:2003:CONLL}.
PER is the easiest type for all systems. Even name matching, 
without any learning, can correctly predict in half of the cases. 

For LOC, it turns out that candidate selection is a bottleneck: when 
candidate selection was flawless, 
the models made only about 12\% errors, 
down from about 57\%. For MISC a similar conclusion can be drawn.

\begin{table*}[!ht]
    \centering
    \begin{tabular}{c|c|c|c|c|c|c|c|c}
         & \multicolumn{4}{c|}{All} & \multicolumn{4}{c}{In $E^+$} \\
         \hline
        System & LOC & ORG & PER & MISC & LOC & ORG & PER & MISC \\
        \hline 
        Name matching & 96.26 & 89.48 & 57.38 & 96.60 & 92.32 & 76.87 & 47.40 & 76.29 \\
        MIL & 57.09 & $\mathbf{76.30}$ & 41.35 & 93.35 & 11.90 & $\mathbf{47.90}$ & 27.60 & 53.61 \\
        MIL-ND & 57.15 & 77.15 & 35.95 & 92.47 & 12.02 & 49.77 & 20.94 & 47.42 \\
        $\tau$MIL-ND & $\mathbf{55.15}$ & 76.56 & $\mathbf{34.03}$ & $\mathbf{92.15}$ & $\mathbf{11.14}$ & 51.18 & $\mathbf{20.59}$ & $\mathbf{40.00}$ \\
        \hline
        Supervised learning &  55.58 & 61.32 & 24.98 & 89.96 & 8.80 & 14.95 & 7.40 & 29.90 \\
    \end{tabular}
    \caption{\% errors on the development set for different named entity types under two settings. (Smaller is better.)}
    \label{tab:ner}
\end{table*}

\begin{figure}
    \centering
    \includegraphics[width=0.48\textwidth]{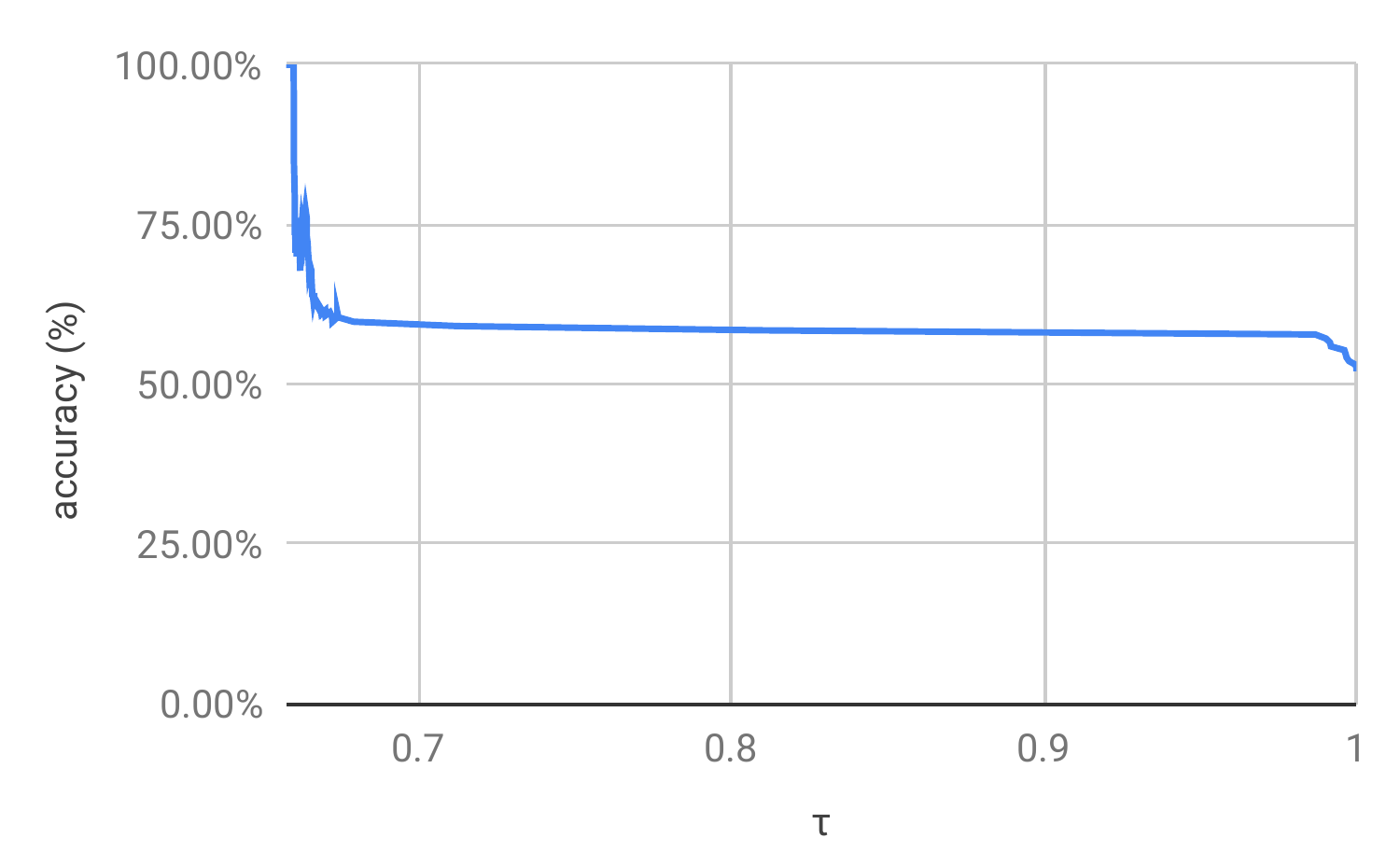}
    \caption{Accuracy vs $\tau$. There are large plateaus between 
    $\tau \in (0.7, 0.95)$ and $\tau < 0.6$ because 
    the ND classifier hardly used these ranges. We hence set $\tau=0.75$ 
    in $\tau$MIL-ND.}
    \label{fig:acc_vs_tau}
\end{figure}

\paragraph{Can the ND classifier detect noise?}
From the training set, we collected 100 data points and manually checked if 
a data point is valid (i.e., $E^+$ contains the correct entity). 
We then checked how the accuracy changes depending on the threshold $\tau$ (Figure~\ref{fig:acc_vs_tau}), the accuracy is defined as
\[
\frac{\text{\# valid data points with } p_N < \tau}{\text{\# all data points with } p_N < \tau}
\]
As expected, the smaller $\tau$ is, 
the higher the chance is that the chosen data point is valid (i.e., not noise). 
Hence, we can use the ND classifier to select high quality 
data points by adjusting $\tau$.

For a further examination, 
from the training set, we collected all 47,213 data points (i.e. 27.8\%)
with $p_N(1|m, c, E^+) > \tau=0.75$, and randomly chose 100 data points. 
We found that 89\% are indeed noisy. This further confirms that the ND classifier
is sufficiently accurate.
Some examples are given in Table~\ref{tab:ip_examples}.

\begin{table*}
    \centering
    \begin{tabular}{|p{0.95\textwidth}|}
    \hline
Correctly detected as noise:

* Small-market teams , like Milwaukee and San Diego , even big-market clubs , like Boston , Atlanta and the two [Chicago] franchises , trumpet them .

Candidates: \textsc{Chicago} (music single), \textsc{Boston to Chicago} (music track)\\

* The politically powerful [Green] movement in Germany has led public opposition to genetic technology research and production .
 
Candidates: \textsc{The Green Prince} (movie), \textsc{The Green Archer} (movie), \textsc{Green Gold} (movie)\\
        \hline 
Incorrectly detected as noise:

* Everything Forrest remains unaffected by , [Jenny] self-indulgently , self-destructively drowns in : radical politics , drug abuse , promiscuity .

Candidates: \textsc{Jenny Curran} (book/film character) \\
    \hline
    \end{tabular}
    \caption{Examples of 100 randomly chosen sentences from the training set 
    whose $p_N(1|m, c, E^+) > \tau=0.75$. The first two examples are correctly 
    detected as noise by our ND classifier. The last one is incorrectly
    detected.}
    \label{tab:ip_examples}
\end{table*}

\paragraph{Number of positive candidates}
\begin{figure}[!ht]
    \centering
    \includegraphics[width=0.48\textwidth]{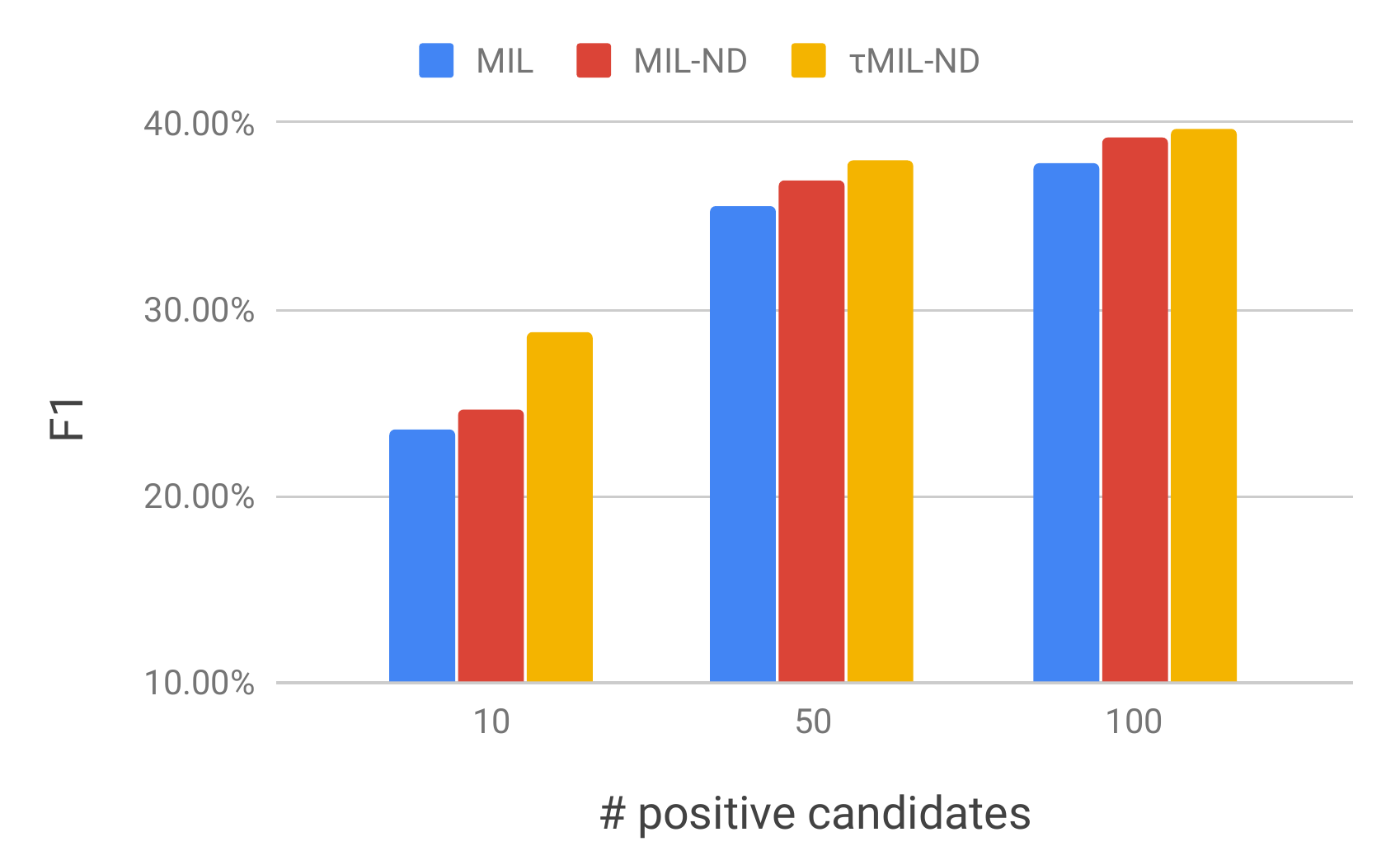}
    \includegraphics[width=0.48\textwidth]{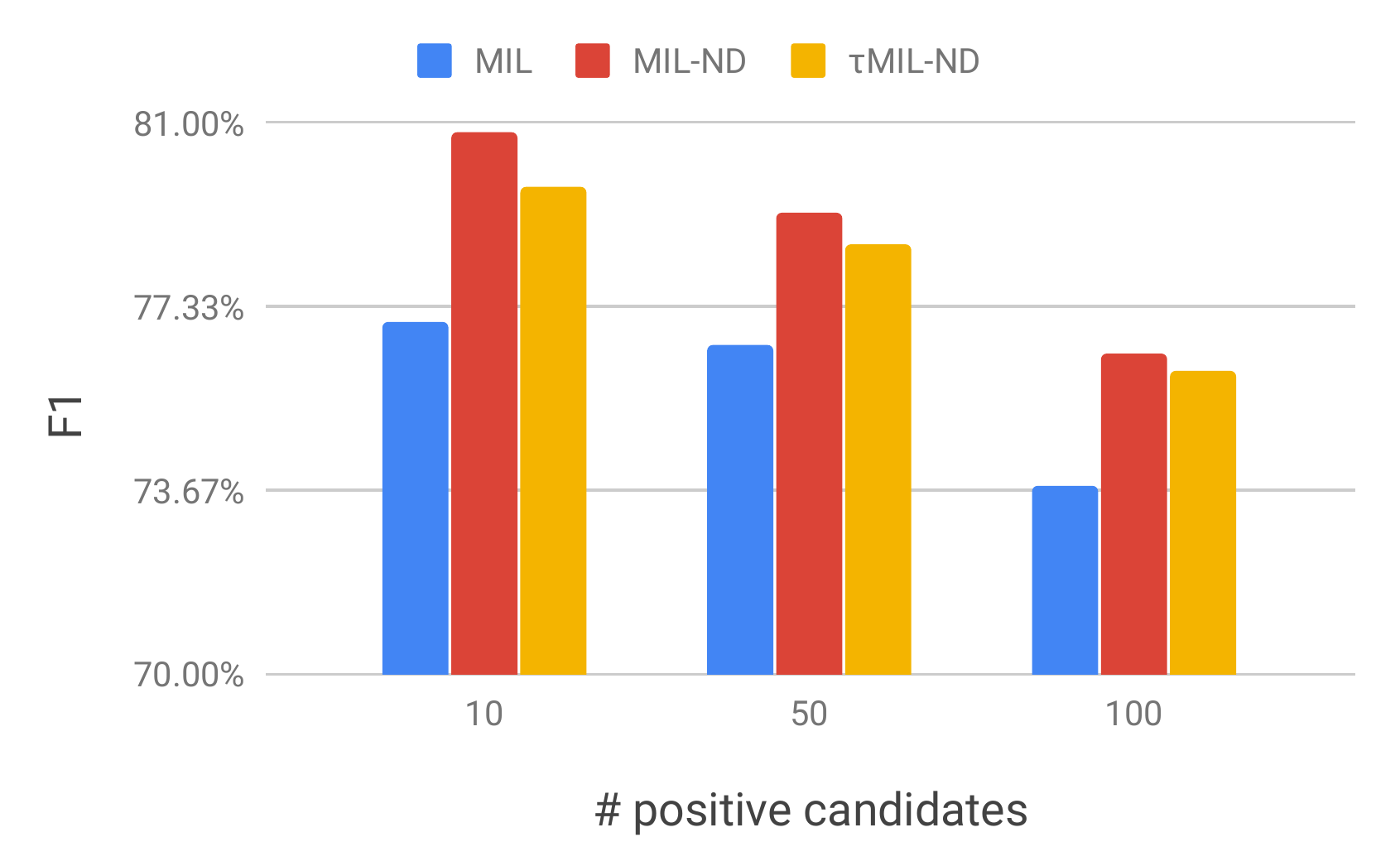}
    \caption{F1 (top:`All'; bottom:`In $E^+$') on the development set with different numbers of 
    positive candidates.}
    \label{fig:f1_n_poss}
\end{figure}

We also experimented with different values of $|E^+|$ (10, 50, 100)
on the development set (Figure~\ref{fig:f1_n_poss}).


First, MIL-ND and $\tau$MIL-ND are always better than MIL. This is more apparent in the `In $E^+$' settings: with this evaluation regime, we zoom in on cases where our models can predict correct entities (of course, all models equally fail for examples outside $E^+$). 

Using the ND classifier at test time to decide to predict any entity or skip ($\tau$MIL-ND) is helpful in the more realistic `All' setting. The difference between $\tau$MIL-ND and MIL-ND is less pronounced for larger $E^+$. This is expected as the proportion of valid data points is higher, and hence the ND classifier is less necessary at test time. 
For `in $E^+$' setting, $\tau$MIL-ND performs worse than MIL-ND, 
as we expected, because there are no noisy data points at test time.


\paragraph{What is wrong with the candidate selector?}
The above results show that candidate selection is a bottleneck and that the used selector is far from perfect. 
We found two cases where the selector is problematic: 
(i) the mention or the entity name is in an abbreviated form, such as 
    `U.N.' rather than `United Nations', 
(ii) the mention and the entity's name only fuzzily match, such as `[English] county' 
    and $\textsc{England}$ (country).
We can overcome these problems via extending our surface matching 
as in \newcite{CHARTON14.899,usbeck2014agdistis} or using word embeddings. 

Even in some cases when the selector does not have any problems 
with surface matching, the number of candidates may be too large. 
For instance, consider `[Simpson] killed his wife...', 
there are more than 1,500 entities in the knowledge base containing the word `Simpson'. It is unlikely that our entity disambiguation model can deal with such 
large lists. We may need a stronger mechanism 
for reducing the number of candidates. For example, we could use document-level information 
to discard highly unlikely entities.

\section{Related work}
\label{sec:related work}


High performance approaches to EL, such as \newcite{P11-1138, 
TACL494, globerson-EtAl:2016:P16-1, TACL1065, ganea-hofmann:2017:EMNLP2017}, 
are two-stage methods: candidate generation is followed by selecting an entity for the candidate lists.
We follow the same paradigm but with some important differences discussed below.

Most approaches use alias-entity maps, i.e.
weighted sets of (mention, entity) pairs 
created from anchors in Wikpedia. For example, one can count
how many times phrase ``the president''
refers to \textsc{Bill Clinton} to assign the weight to the corresponding pair.
However,
the method requires large annotated datasets, and it cannot deal with less prominent entities.
As we do not have access to links, we use surface matching instead. 

To choose an entity from a 
candidate list, 
two main disambiguation frameworks \cite{P11-1138}
are introduced: {\it local} which 
resolves mentions independently, and {\it global}
which makes use of coherence modeling at the document level. 
Though we experimented with local models, the local-global distinction is largely orthogonal as we can directly integrate coherence modeling components in our DL approach.
%

Different types of supervision have been considered in previous work: 
full supervision \cite{TACL1065, ganea-hofmann:2017:EMNLP2017, P18-1148}, 
using combinations of labeled and unlabeled data \cite{TACL637}, 
and even distant supervision \cite{Y15-1010}. 
The approach of \newcite{Y15-1010}
is heavily Wikipedia-based: 
they rely on a heuristic mapping from Freebase entities to Wikipedia entities, 
and learn features from Wikipedia articles. Unlike ours, their approach cannot be generalized to  set-ups where no documents are available for entities. 



\section{Conclusions}
\label{sec:conclusion}

We introduced the first approach to entity linking which neither uses annotated texts, nor assumes that entities are associated with textual documents (e.g., Wikipedia articles). 

We learn the model using the MIL paradigm, and introduce a novel component, a noise detecting classifier, estimated jointly with the EL model. The classifier lets us disregard noisy labels, resulting in a more accurate entity linking model.
Experimental results showed that our models substantially outperform the heuristic baseline, and, for certain categories, they approach the model estimated with supervised learning.

In future work we will aim to improve candidate selection (including 
different strategies to select candidate lists $E^+, E^-$).
We will also use extra document information and jointly predict entities for different mentions in the document. Besides, we will consider additional knowledge bases (e.g. YAGO and Wikidata).

\section*{Acknowledgments}
We would like to thank anonymous reviewers for their 
suggestions and comments. The project was supported by the
European Research Council (ERC StG BroadSem
678254), the Dutch National Science Foundation
(NWO VIDI 639.022.518), and an Amazon Web Services
(AWS) grant.

\bibliography{ref}
\bibliographystyle{acl_natbib}

\appendix

\section{Hyper-parameters}
\label{app:hp}
The used values for the hyper-parameters are shown in 
Table~\ref{tab:hyper-parameters}.

\begin{table}[!ht]
\centering
\begin{tabular}{l|c|c}
Hyper-parameters & Model & Value \\
\hline

learning rate (Adam) & 1, 2 & 0.001 \\
mini-batch size & 1, 2 & 50 \\
number of epochs & 1, 2 & 20 \\
$d_w$ (word emb. dim.) & 1, 2& 300 \\
$d_p$ (position emb. dim.) & 1, 2& 5 \\
$d_t$ (type emb. dim.) & 1, 2& 50 \\
$d_e$ (entity emb. dim.) & 1, 2& 100 \\
BiLSTM hidden dim. & 1, 2& 100 \\
$\text{FFN}_{g}$ hidden dim. & 1, 2& 300 \\
$\text{FFN}_{h}$ hidden dim. & 2& 300 \\
$\delta$ (margin) & 1, 2& 0.1 \\
$T$ (temperature) & 1, 2& $1/3$ \\
$\eta$ (KL coefficient) & 2& 5 \\
$p^*_I(1)$ (prior) & 2 & 0.9 \\
$\tau$ (threshold) & 2 & 0.75 \\
\end{tabular}
\caption{Values of hyper-parameters.}
\label{tab:hyper-parameters}
\end{table}

\section{Supervised learning system}
Our supervised learning system is a supervised version of 
model 1. To examine how good this system is, we tested it 
on the AIDA CoNLL dataset. 
Because this system uses entity types for entity disambiguation
(and uses no other information related to entities), 
we made use of the map between Freebase entities and Wikipedia entities. 
We compared it with \newcite{hoffart-EtAl:2011:EMNLP}. Note that 
we did not tune the system: it used the same values 
of hyper-parameters with model 1 (see Table\ref{tab:hyper-parameters}).

Because Wikipedia and YAGO are often used for candidate 
selection and/or for additional supervision, we here also used them
for candidate selection and for computing $p(e|m)$ as a feature 
as in most existing systems (such as in \newcite{hoffart-EtAl:2011:EMNLP, globerson-EtAl:2016:P16-1, ganea-hofmann:2017:EMNLP2017}). 
For the candidate section, for each mention 
we kept maximally 20 candidates.

We ran our system five times and report mean and 95\% confidence interval.
Table~\ref{tab:sup} shows micro accuracy (in knowledge-base).
Our system performs on par with \newcite{hoffart-EtAl:2011:EMNLP}.

\begin{table}
    \centering
    \begin{tabular}{l|c}
        System & Micro accuracy \\  
        \hline
        Ours & 81.47 $\pm 1.27$ \\
        \newcite{hoffart-EtAl:2011:EMNLP} & 81.91
    \end{tabular}
    \caption{Micro accuracy on AIDA CoNLL testb 
    of our supervised system and \cite{hoffart-EtAl:2011:EMNLP}.}
    \label{tab:sup}
\end{table}

\end{document}